\documentclass[10pt,twocolumn,letterpaper]{article}

\usepackage[pagenumbers]{wacv} 

\usepackage{multirow}
\usepackage{colortbl}
\usepackage{xcolor}

\definecolor{wacvblue}{rgb}{0.21,0.49,0.74}
\usepackage[pagebackref,breaklinks,colorlinks,allcolors=wacvblue]{hyperref}

\def\wacvPaperID{1853} 
\def\confName{WACV}
\def\confYear{2026}

\title{SAVE: Sparse Autoencoder‑Driven Visual Information Enhancement \\for Mitigating Object Hallucination}

\author{
  Sangha Park$^1$ \hspace{0.2em}
  Seungryong Yoo$^1$ \hspace{0.2em}
  Jisoo Mok$^2$\textsuperscript{†}\hspace{0.2em}
  Sungroh Yoon$^{1,3}$\textsuperscript{†}
  \vspace{0.3em} \\
  $^1$Department of Electrical and Computer Engineering, Seoul National University \hspace{0.2em} \\
  $^2$Daegu Gyeongbuk Institute of Science and Technology\\
  $^3$IPAI, AIIS, ASRI, INMC, and ISRC, Seoul National University \\
}

\begin{document}
\maketitle
 \begin{abstract}
Although Multimodal Large Language Models (MLLMs) have advanced substantially, they remain vulnerable to object hallucination caused by language priors and visual information loss.
To address this, we propose \textbf{SAVE} (\textbf{S}parse \textbf{A}utoencoder‑Driven \textbf{V}isual Information \textbf{E}nhancement), a framework that mitigates hallucination by steering the model along Sparse Autoencoder (SAE) latent features.
A binary object‑presence question‑answering probe identifies the SAE features most indicative of the model’s visual information processing, referred to as \textbf{visual understanding features}.
Steering the model along these identified features reinforces grounded visual understanding and effectively reduces hallucination.
With its simple design, SAVE outperforms state‑of‑the‑art training‑free methods on standard benchmarks, achieving a 10\%p improvement in $\text{CHAIR}_S$ and consistent gains on POPE and MMHal‑Bench.
Extensive evaluations across multiple models and layers confirm the robustness and generalizability of our approach.
Further analysis reveals that steering along visual understanding features suppresses the generation of uncertain object tokens and increases attention to image tokens, mitigating hallucination. Code is released at \url{https://github.com/wiarae/SAVE}.
\end{abstract}    
\let\thefootnote\relax\footnotetext{†Corresponding authors}
\section{Introduction}

Multimodal Large Language Models (MLLMs) have achieved strong performance across various multimodal tasks such as image captioning, visual question answering, and multimodal reasoning by aligning visual inputs with pretrained language models~\cite{dai2023instructblip, liu2023visual}.
However, these models are prone to object hallucination—the generation of text describing non-existent objects or attributes not grounded in the visual input~\cite{bai2024hallucination, leng2024mitigating}.
Unlike hallucination in unimodal LLMs, which mainly arises from linguistic priors, hallucination in MLLMs additionally involves failures in visual grounding, stemming from misaligned attention or suppressed visual cues~\cite{huang2024opera, liu2024paying}.

\begin{figure}[t]
  \begin{center}
\includegraphics[width=0.85\columnwidth]{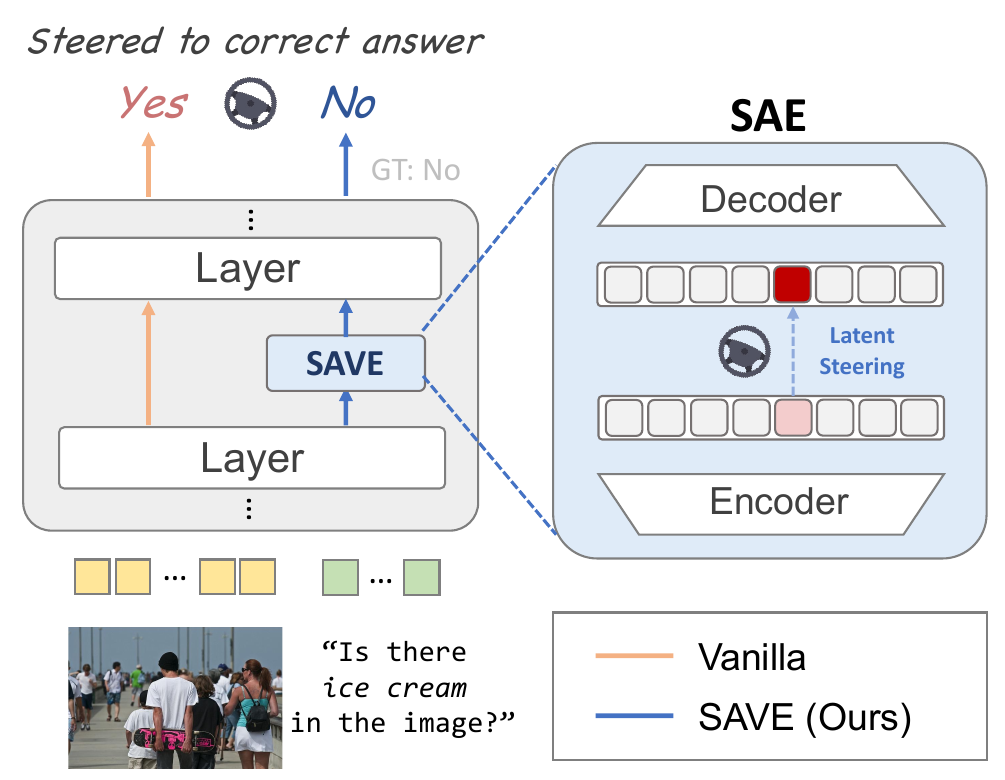}
  \end{center}
\vskip -0.2in
   \caption{Illustration of SAVE. 
   While vanilla model produces a hallucinated response, our approach—using SAE latent steering—successfully generates a correct, grounded response.
   }
   \label{fig:teaser_image}
\end{figure}
To mitigate object hallucination, prior studies have focused on reducing language priors and alleviating visual information loss through strategies such as contrastive decoding~\cite{leng2024mitigating, suo2025octopus} and latent steering~\cite{li2025hiddenlifetokensreducing, liu2025reducing}.
More recently, studies have observed that the internal behavior of MLLMs differs between responses containing hallucinated objects and those correctly describing ground‑truth objects~\cite{jiang2025interpreting, jiang2025devils}, suggesting that hallucination arises from a structured shift in internal dynamics rather than random failure.
These insights motivate approaches aimed at strengthening the model’s visual information processing,  which first calls for a clear understanding of the model’s visual encoding capabilities.

Sparse autoencoders (SAEs), widely used in mechanistic interpretability~\cite{unknown-author-no-date}, decompose model activations into sparse and often monosemantic features, where each latent dimension corresponds to a semantically meaningful direction~\cite{templeton2024scaling, bricken-2023}.
Originally proposed for LLMs~\cite{lieberum2024gemma}, SAEs have recently been extended to multimodal models~\cite{zhang2024largemultimodalmodelsinterpret, rajaram2025line}.
Building on these efforts, we leverage SAEs to capture visual information processing of MLLMs and enable targeted steering that reinforces grounded visual understanding.

Surprisingly, we demonstrate that SAEs, when guided by a well‑designed probe, provide an effective approach for capturing a model’s visual information processing.
We formulate binary object‑presence question‑answering as such a probe, enabling the computation of a separation score that measures differences in SAE activations between correct and hallucinated responses.
SAE features with the highest separation scores are identified as \textbf{visual understanding features}, serving as the primary latent directions of the model’s visual information processing, while features with high separation in the opposite direction are considered hallucinated features.
Our analysis further reveals that these SAE features are semantically disentangled: visual understanding features predominantly activate on correct responses, whereas hallucinated features are more frequent in hallucinated ones.
This selective activation faithfully reflects the model’s visual information processing and reveals its underlying failure modes.

Building on these insights, we propose \textbf{SAVE} (\textbf{S}parse \textbf{A}utoencoder‑Driven \textbf{V}isual Information \textbf{E}nhancement), which mitigates object hallucination by steering along visual understanding features.
Reinforcing these features effectively reduces hallucination, whereas steering along hallucinated features amplifies it.
Although derived from an object‑level probe, the identified features reliably mitigate hallucination through latent steering, revealing the model’s susceptibility to visual information loss and language priors—the key drivers of object hallucination.

We evaluate SAVE on POPE~\cite{li-etal-2023-evaluating}, CHAIR~\cite{rohrbach-etal-2018-object}, and MMHal‑Bench~\cite{sun-etal-2024-aligning}.
With its simple design, SAVE consistently outperforms multiple state‑of‑the‑art training‑free approaches.
In particular, compared to their respective base models, CHAIR\textsubscript{S} improves by about 10\%p on LLaVA‑1.6~\cite{liu2024improved}, 5\%p on LLaVA‑NeXT~\cite{liu2024llavanext}, and 20\%p on Qwen2-VL~\cite{wang2024qwen2}.
Moreover, experiments show that applying steering at different layer positions effectively mitigates hallucination, demonstrating our approach is layer‑wise applicable.
By validating on both LLaVA models and Qwen2‑VL~\cite{wang2024qwen2}, we demonstrate that our method is robust and generalizable across MLLMs.
Finally, our analysis shows that steering along visual understanding features mitigates hallucination by lowering the probability of uncertain object tokens and increasing attention to image tokens.

Our contributions are as follows:
\begin{enumerate}[itemsep=1pt, parsep=0pt]
\item We propose SAVE, a novel framework that leverages SAE features to interpret a model’s visual information processing through binary object‑presence probing, thereby enabling effective hallucination mitigation.
\item We demonstrate that steering along discovered visual understanding features significantly reduces hallucination and consistently outperforms state‑of‑the‑art training‑free methods.
\item Through extensive experiments across multiple MLLMs (LLaVA‑1.6, LLaVA‑NeXT, Qwen2‑VL) and layer‑wise steering, we demonstrate the robustness and generalizability of our SAE‑based approach.

\end{enumerate}

\section{Related work}
\subsection{Object Hallucination in MLLMs
}

Object hallucination—where models generate content that is factually incorrect, irrelevant, or inconsistent with visual inputs~\cite{bai2024hallucination}—is a persistent challenge in MLLMs. Recent studies address this issue through three main approaches. Decoding-based methods (e.g., VCD~\cite{leng2024mitigating}, SID~\cite{huo2025selfintrospective}, DeCo~\cite{wang2025mllm}, ED~\cite{choyou}) enforce token-level grounding by adjusting the logits during generation. Attention-based approaches (e.g., PAI~\cite{liu2024paying}, VAR~\cite{kang2025see}, AD-HH~\cite{yangunderstanding}) refine attention maps to improve image-text alignment. Steering-based methods (e.g., VTI~\cite{liu2025reducing}, VISTA~\cite{li2025hiddenlifetokensreducing}) manipulate internal activations to suppress hallucinated content.
These efforts suggest hallucination is shaped by controllable internal dynamics. We build on this view by introducing SAEs for representation-level analysis and control of hallucination in multimodal models.

\begin{figure}[t]
  \centering
  \includegraphics[width=\columnwidth]{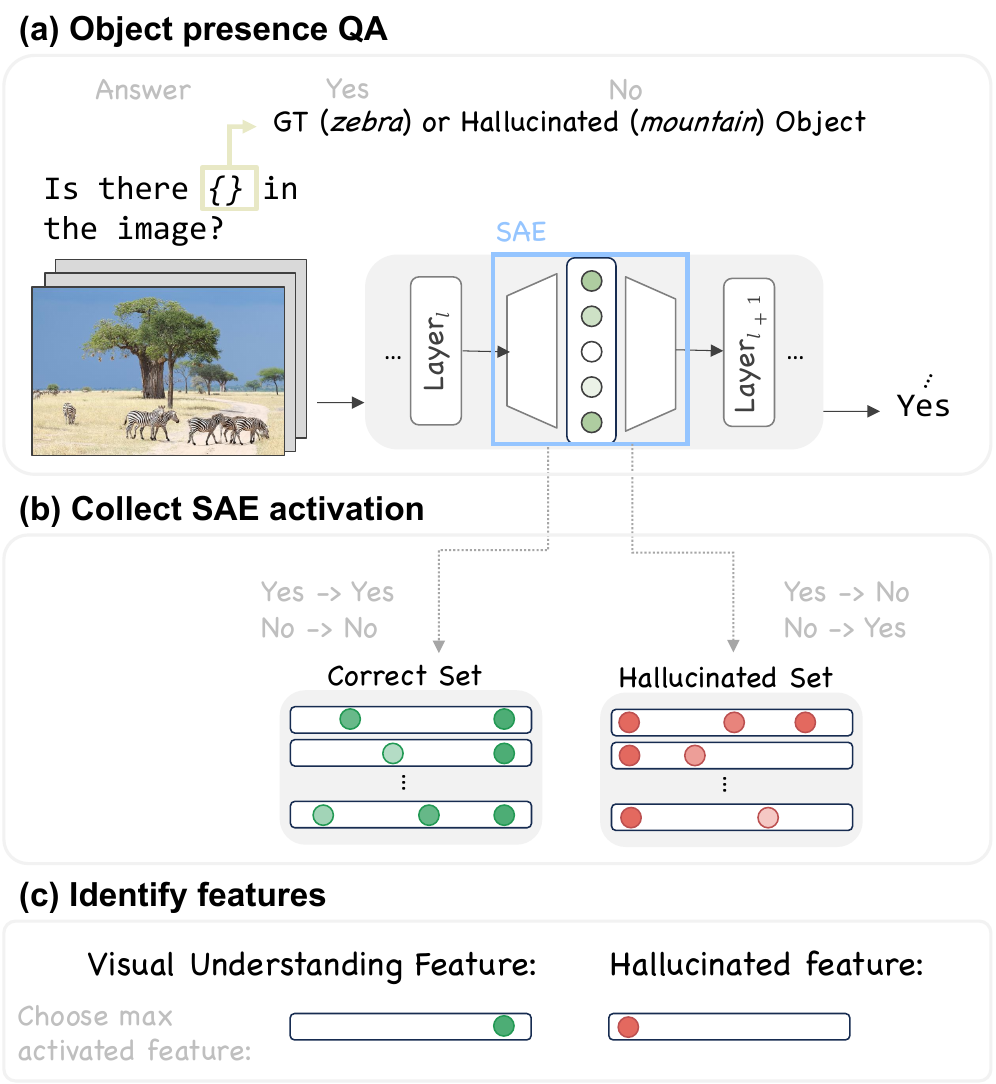}
  \caption{Overall framework of SAVE. (a) We obtain SAE activations using object-presence yes/no questions, then (b) split them by response correctness. (c) Separation scores are computed to identify features associated with visual understanding (correct cases) and hallucination (incorrect cases).}\label{fig:method}
\end{figure}

\subsection{Mechanistic Interpretability with SAE}
Mechanistic interpretability~\cite{unknown-author-no-date} aims to uncover how neural networks produce their outputs by analyzing internal activations and identifying human-interpretable features associated with individual neurons~\cite{olah2020zoom, elhage2022toymodelssuperposition}. However, analyzing individual neurons is challenging due to their polysemantic nature—a phenomenon caused by superposition, where models encode more features than available dimensions.~\cite{elhage2022toymodelssuperposition}.
To address the polysemanticity of neurons, SAEs decompose activations into a sparse set of latent directions that form a high-dimensional feature dictionary~\cite{bricken-2023}. SAEs have been used to discover interpretable and steerable features of LLMs, including those related to bias and safety~\cite{cunningham2023sparse, templeton2024scaling}. 

Building on this, several studies have applied SAEs to vision-language models such as CLIP, enabling visual concept extraction and image generation steering~\cite{daujotas-2024}. More recently, SAEs have been extended to MLLMs, where they are used to analyze visual concepts~\cite{lim2025sparse}, track the emergence of multimodal alignment~\cite{lou2025saevinterpretingmultimodalmodels}, and identify semantically meaningful features across modalities~\cite{zhang2024largemultimodalmodelsinterpret}.
This paper extends these efforts by utilizing SAEs for hallucination control in MLLMs, leveraging their interpretability to analyze and steer internal representations.

\section{Preliminaries}

\subsection{Sparse Autoencoder}
SAEs are a powerful tool for discovering interpretable structure in model activations. Inspired by classical dictionary learning~\cite{olshausen1997sparse}, they learn a sparse representation of the input by constraining the intermediate activations such that only a small number of latent units are active at a time. 

To enforce this sparsity, the SAE architecture proposed by OpenAI~\cite{gao2025scaling} introduces a two-layer autoencoder with a TopK activation function, which retains only the top-$k$ activations and zeroes out the rest. Given a model representation $x$, the SAE defines the encoding and decoding process as:
\begin{equation}\label{eq:sae}
a(x) = \mathrm{TopK} \left( \mathrm{ReLU}(W_{\mathrm{enc}} (x - b_{\mathrm{pre}}) + b_{\mathrm{enc}}) \right),
\end{equation}
\begin{equation}\label{eq:sae_dec}
\mathrm{SAE}(x) = W_{\mathrm{dec}} \cdot a(x) + b_{\mathrm{dec}}.
\end{equation}
Here, $W_{\mathrm{enc}}$ and $b_{\mathrm{enc}}$ denote the encoder’s weight matrix and bias, while $W_{\mathrm{dec}}$ and $b_{\mathrm{dec}}$ are the corresponding decoder parameters.
A learned normalization offset $b_{\mathrm{pre}}$ is applied to the input $x$ prior to encoding. We denote the input as $x \in \mathbb{R}^{T \times d_n}$, where $T$ is the number of tokens and $d_n$ is the hidden dimension of the model.
The encoder produces a sparse activation $a(x) \in \mathbb{R}^{T \times d_M}$, where the latent dimension $d_M$ is much larger than the input dimension $d_n$ (i.e., $d_M \gg d_n$). This allows the input to be represented over a wide dictionary of features.
We refer to the sparse output $a(x)$ as the \textit{latent activation}, and to each row of $W_{\mathrm{dec}}$ as a \textit{latent direction}, which serves as a basis vector in the learned dictionary.
Each dimension of the $d_M$-dimensional latent space is indexed by $j \in \{1, \dots, d_M\}$, and we refer to each index $j$ as a distinct \textit{latent feature}, representing a specific decoder direction.
For a given input $x$, the activation value $a_{t,j}(x)$ denotes the strength of feature $j$ at token position $t$.



SAE is trained to minimize the following loss function:
\begin{equation}
\mathcal{L}(x) =
\underbrace{\left\| x - \mathrm{SAE}(x) \right\|_2^2}_{\mathcal{L}_{\text{reconstruction}}}
+ \lambda \underbrace{\left\| a(x) \right\|_0}_{\mathcal{L}_{\text{sparsity}}}.
\end{equation}
where the first term encourages accurate reconstruction of the input, and the second imposes an $\ell_0$-based sparsity penalty on the activation vector $a(x)$. The hyperparameter $\lambda$ controls the balance between these objectives.
This training setup reconstructs the input as a sparse linear combination of latent directions, which tend to represent human-interpretable features.  The sparsity constraint promotes monosemanticity, where each latent unit captures a distinct, coherent concept~\cite{cunningham2023sparse, bricken-2023, templeton2024scaling, gao2025scaling}.




\begin{figure*}[t]
  \centering
  \includegraphics[width=\textwidth]{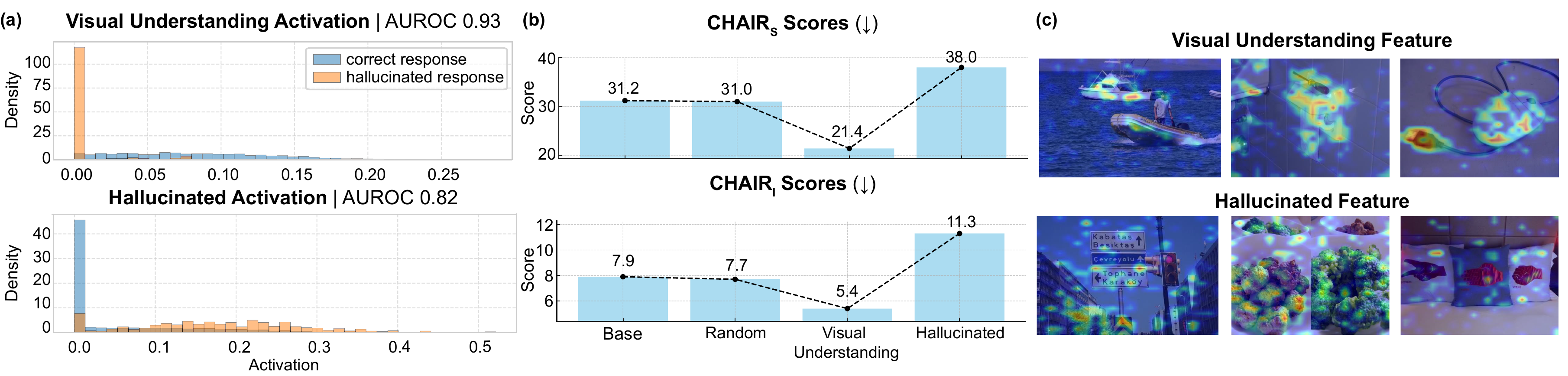}
  \caption{Analysis of identified features. (a) AUROC of visual‑understanding and hallucinated features, computed by activation analysis on correct versus hallucinated responses. (b) CHAIR evaluation of steering behavior toward random, visual‑understanding, and hallucinated features. (c) Visualization of each feature using its top‑activated images and the most responsive patches.}
  \label{fig:VU_HU}
\end{figure*}

\subsection{Multimodal SAE}\label{sec:multimodal-sae}
Recent work extends sparse autoencoders (SAEs) from text-only LLMs to large multimodal models, showing that SAEs can disentangle cross-modal, open-semantic features and even steer model behavior (e.g., feature-level interventions and transfer across model scales)~\cite{lim2025sparse, zhang2024largemultimodalmodelsinterpret}. Building on this motivation, the multimodal‑SAE~\cite{rajaram2025line} is integrated into intermediate transformer layers of LLaVA‑1.6-Mistral-7B to decompose residual stream activations into sparse, semantically meaningful features.
Following prior work on scaling SAEs~\cite{templeton2024scaling, gao2025scaling}, a ReLU‑based encoder–decoder architecture with an 8× expansion ratio is employed and trained to minimize a combination of reconstruction loss and sparsity regularization.

The SAE learns a dictionary of 32k latent features, with sparse activations that promote disentangled visual‑language representations.
It is trained on roughly 1.2M image–caption pairs from ShareGPT4V~\cite{chen2024sharegpt4v}.
To capture cross‑layer visual grounding behavior, SAEs are trained at layers 8, 12, 16, 20, and 24.
Additional details on density scheduling, token counts, and optimization are provided in~\Cref{sec:appen_conf_details}.
\section{Visual Understanding Feature}\label{sec:detect}
Object hallucination can be mitigated by enhancing visual information of the model~\cite{suo2025octopus, leng2024mitigating}.
Prior studies have shown that the model’s internal representations are distinct when producing ground-truth versus hallucinated tokens~\cite{jiang2025interpreting, jiang2025devils}.
Building on these insights, we leverage the disentangled nature of SAEs to identify features most relevant to visual grounding and steer the model along these directions.
This targeted steering enhances the model’s visual information, guiding it toward faithful object descriptions and reducing hallucinations.

\begin{figure}[t]
  \begin{center}
\includegraphics[width=\columnwidth]{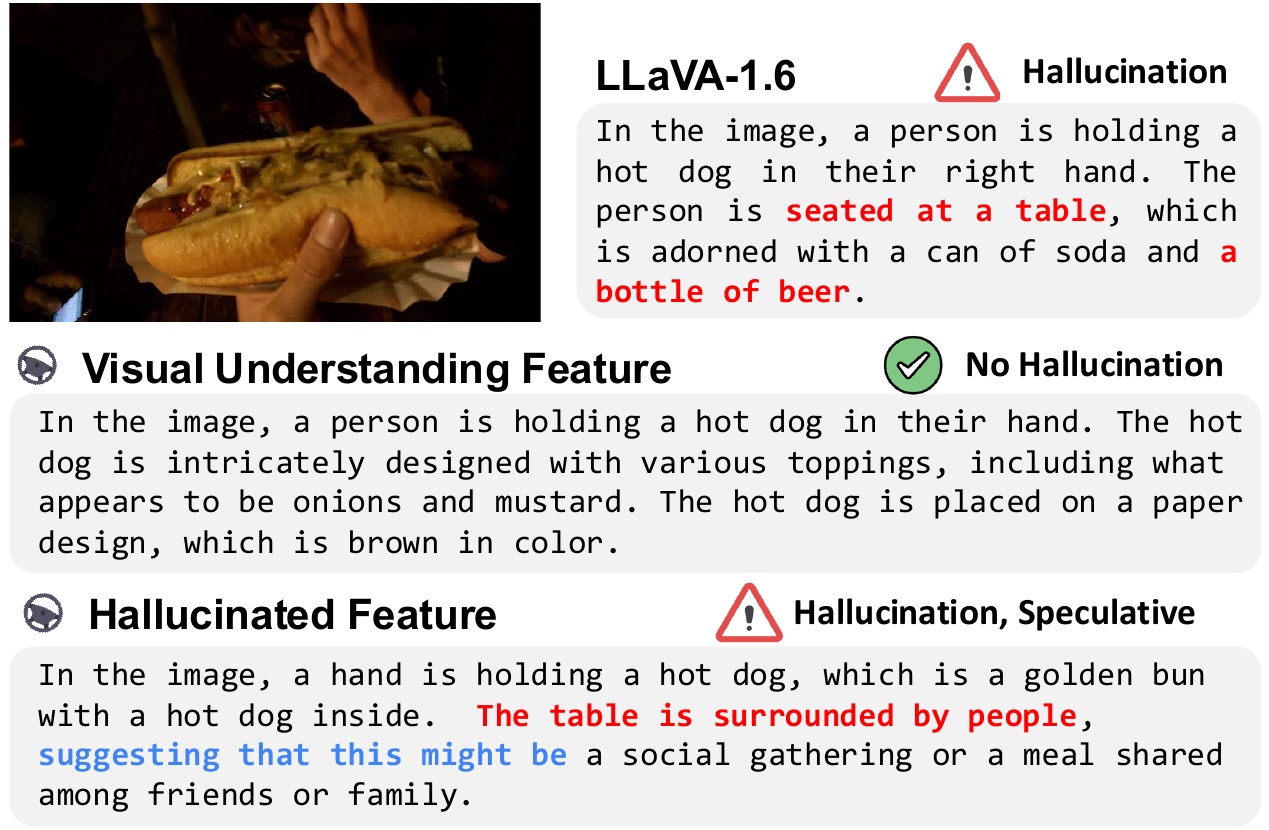}
  \end{center}
   \caption{Qualitative examples of model steering. Steering toward visual understanding features mitigates object hallucination, whereas steering toward hallucinated features yields responses that remain hallucinated (highlighted in red) and include speculative wording (highlighted in blue).}
   \label{fig:chair}
\end{figure}

\begin{table*}[t]
\centering
\resizebox{\textwidth}{!}{%
\begin{tabular}{ll|cc|cc|cc|cc|cc|cc}
\toprule
& & 
\multicolumn{2}{c|}{\multirow{2}{*}{\centering \textbf{CHAIR}}} &
\multicolumn{6}{c|}{\textbf{POPE}} &
\multicolumn{2}{c}{\multirow{2}{*}{\centering \textbf{MMHal-Bench}}} \\
& & 
& & 
\multicolumn{2}{c|}{Random} 
& \multicolumn{2}{c|}{Popular} 
& \multicolumn{2}{c|}{Adversarial} 
& & \\
\cmidrule(r){3-4} \cmidrule(r){5-6} \cmidrule(r){7-8} \cmidrule(r){9-10} \cmidrule(r){11-12}
& & 
\textbf{CHAIR\textsubscript{S}}$\downarrow$ & \textbf{CHAIR\textsubscript{I}}$\downarrow$ & 
\textbf{F1} $\uparrow$& \textbf{Acc} $\uparrow$& 
\textbf{F1} $\uparrow$& \textbf{Acc} $\uparrow$& 
\textbf{F1} $\uparrow$& \textbf{Acc} $\uparrow$& 
\textbf{Score}$\uparrow$ & \textbf{HalRate}$\downarrow$ \\
\midrule
\multirow{7}{*}{\shortstack{\textbf{LLaVA-}\\\textbf{1.6-7B}}}
&Base           & 31.2 & 7.9  & 92.16 & \underline{92.19} & 89.66 & 89.73 & 86.32 & 85.90 & 2.88 & \underline{0.41} \\
&VCD            & 38.6 & 10.1 & 90.01 & 89.93 & 87.27 & 87.23 & 83.72 & 82.96 & 2.65 & 0.48 \\
&DeCO           & 38.4 & 11.7 & \underline{92.32} & 91.05 & 88.79 & 88.16 & 83.55 & 81.53 & 2.64 & 0.47 \\
&Devils         & 36.0 & 9.8  & 90.19 & 90.48 & 89.49 & \textbf{90.03} & \textbf{86.60} & \textbf{86.86} & 2.25 & 0.54 \\
&VTI            & 29.4 & \underline{5.8}  & 91.78 & 91.75 & 88.91 & 89.48 & 84.46 & 84.97 & 2.54 & 0.47 \\
&VISTA          & \underline{26.8} & 11.3 & 91.93 & 91.99 & \underline{89.71} & \underline{89.80} & 86.28 & 85.86 & \underline{2.93} & \underline{0.41} \\
&\textbf{SAVE (Ours)} & \textbf{21.4} & \textbf{5.4}  & \textbf{92.71} & \textbf{92.61} & \textbf{89.74} & 89.77 & \underline{86.52} & \underline{86.37} & \textbf{3.12} & \textbf{0.36} \\
\midrule
\multirow{7}{*}{\shortstack{\textbf{LLaVA-}\\\textbf{NeXT-8B}}}
& Base          & 34.2 & 9.0  & 92.74 & \underline{92.65} & 89.88 & \underline{89.73} & \underline{85.60} & 84.67 &  2.83 & 0.45 \\
& VCD           & 35.4 & 9.7  & 90.53 & 90.44 & 88.16 & 88.00 & 84.98 & 84.20 & 2.90  & 0.45   \\
& DeCO  & 35.6 & 11.0  & \underline{92.75} & 92.61 & \underline{89.72} & 89.50  & 85.38      & 84.30 & 2.75  & 0.46  \\
& Devils  & \textbf{21.4}  & \underline{7.0}  &  90.67  & 90.82  &  88.66  & 88.93      &  85.17 & \textbf{84.93}  &  2.22  & 0.55  \\
& VTI           & 32.8 & \textbf{5.4}  & 91.78 & 91.75 & 89.22 & 89.20 & 83.99 & 82.99 & 2.54 & 0.57  \\
& VISTA         & 33.4 & 8.3  & 92.37 & 92.26 & 89.68 & 89.53 & 85.51 & 84.60 & \underline{3.02} & \underline{0.42}  \\
& \textbf{SAVE (Ours)}  & \underline{28.0} & 7.0  & \textbf{92.85} & \textbf{92.78} & \textbf{89.94} & \textbf{89.83} & \textbf{85.68} & \underline{84.80} & \textbf{3.21} & \textbf{0.29}      \\
\midrule
\multirow{2}{*}{\shortstack{\textbf{Qwen2-}\\\textbf{VL-7B}}}
& Base & 40.0	&7.2 &90.56	&90.89 & 88.17	&88.63		&\textbf{86.43}&	\textbf{86.70} & 3.31 & 0.32\\
& \textbf{SAVE (Ours)} & \textbf{20.2} & \textbf{5.9} & \textbf{92.72} & \textbf{92.68} & \textbf{89.09} & \textbf{88.93} & 85.90 & 85.17 & \textbf{3.70} & \textbf{0.23}\\

\bottomrule
\end{tabular}
}
\caption{Results on CHAIR, POPE, and MMHal‑Bench benchmarks, comparing SAVE with state‑of‑the‑art training‑free methods. Best and second‑best results are highlighted in bold and underlined, respectively. Because most baselines are built on LLaVA, for Qwen2-VL we compare only against the vanilla model. Metric descriptions are provided in \Cref{sec:benchmark}.}
\label{tab:main_results}
\end{table*}

\subsection{Collect SAE Activations}\label{sec:4.1}
SAE features are characterized by how frequently they are activated under specific concepts.
To identify visual understanding features, we introduce a probe for the model’s visual information processing.
Following prior work~\cite{wang2025mllm}, we formulate the probe as a binary object‑presence question‑answering task, in which the model determines whether a queried object is present in the image (see~\Cref{fig:method}).
To avoid biasing the extracted SAE features toward a particular response (e.g., consistently favoring “yes” or “no”), we construct 10,000 balanced queries—5,000 for objects present (GT) and 5,000 for objects absent (hallucinated)—using LURE~\cite{zhou2024analyzing}, a benchmark of hallucinated objects generated by GPT-3.5. Each activation \( a(x) \) (as defined in \Cref{eq:sae}) is assigned to \( \mathcal{X}_{\text{correct}} \) if the response is correct, or \( \mathcal{X}_{\text{hallu}} \) otherwise. Experimental results verifying that there is no bias toward any specific response are provided in~\Cref{sec:analysis}. 

\subsection{Identify Visual Understanding Feature}
While SAE features are typically associated with a concept based on frequent activation under that concept, a more robust identification of visual understanding features can be achieved by explicitly contrasting activations between correct and hallucinated responses~\citep{ferrando2025do}.

For each feature $j$, the activation frequency in each set is computed as
\begin{align}
f^{\text{correct}}_{j} &= \frac{1}{N_{\text{correct}}} \sum_{i=1}^{N_{\text{correct}}} 1\left[a_{j}(x^{\text{correct}}_{i}) > 0 \right], \\
f^{\text{hallu}}_{j}   &= \frac{1}{N_{\text{hallu}}} \sum_{i=1}^{N_{\text{hallu}}} 1\left[a_{j}(x^{\text{hallu}}_{i}) > 0 \right]
\end{align}
where $a_j(x)$ denotes the activation of feature $j$ for input $x$.

A separation score is then defined as
\begin{equation}
s_{j} = f^{\text{correct}}_{j} - f^{\text{hallu}}_{j}
\end{equation}
which captures the degree to which a feature is preferentially activated for visually grounded predictions.
The feature with the highest $s_j$ is selected as the visual understanding feature, i.e., $\arg\max_j s_j$.

Conversely, features strongly associated with hallucination can be identified by reversing the separation score,
\begin{equation}
s_{j} = f^{\text{hallu}}_{j} - f^{\text{correct}}_{j}.
\end{equation}
\paragraph{Activation Analysis} The resulting features exhibit clear discriminative behavior in our experiments: visual understanding features are highly activated for correct responses (AUROC=0.93), while hallucination features are selectively activated for hallucinated responses (AUROC=0.82), as illustrated in~\Cref{fig:VU_HU}(a).

\subsection{Steering Model Behavior}\label{sec:steering}
Adjusting the model’s internal representations along these identified features strengthens visual grounding and mitigates object hallucinations.
Formally, an SAE reconstructs a hidden representation as
\begin{equation}
x \approx \sum_{j} a_{j}(x)\, W_{\text{dec}}[j,:] + b_{\text{dec}},
\end{equation}
which is equivalently expressed as \Cref{eq:sae_dec}, where $a_{j}(x)$ is the activation of feature $j$ and $W_{\text{dec}}[j,:]$ is its decoder direction.
Modulating a single feature $j$ by a scalar $\alpha$ (the steering strength) shifts the representation along its decoder direction, producing
\begin{equation}
x_{\text{steered}} = x + \alpha W_{\text{dec}}[j,:] ,
\label{eq:steering}\end{equation}
which corresponds to activation steering in the latent space~\citep{turner2023activation}.

\paragraph{Model Behavior Analysis} Steering along the visual understanding feature effectively reduces object hallucination, whereas steering along the hallucination feature amplifies it.
As shown in \Cref{fig:VU_HU}(b), CHAIR\textsubscript{S} drops from 31.2 to 21.4 when steering toward the visual understanding feature, while steering toward the hallucination feature increases it to 38.0, with a random baseline obtained by averaging five randomly selected features in between.
Qualitative examples in \Cref{fig:chair} further confirm this trend.

\subsection{Feature Visualization} 
To analyze the semantics of each SAE feature, we visualize the top‑activated images and their most responsive patches.  
Following~\cite{zhang2024largemultimodalmodelsinterpret}, all images are processed through the model to obtain hidden representations 
\( X \in \mathbb{R}^{|D| \times n_\text{img} \times d} \),  
where \( |D| \) is the number of image samples (indexed by \(k\)), \( n_\text{img} \) is the number of image tokens per image (indexed by \(j\)), and \( d \) is the hidden feature dimension.  
These are then projected into the SAE latent space as  
\( Z \in \mathbb{R}^{|D| \times n_\text{img} \times d_s} \),  
where \( d_s \) is the number of SAE latent features (indexed by \(i\)).

We first compute the mean activation of each feature across image tokens:
\[
\bar{Z}[k, i] = \frac{1}{n_\text{img}} \sum_{j=1}^{n_\text{img}} Z[k, j, i],
\quad
\bar{Z} \in \mathbb{R}^{|D| \times d_s},
\]
and select the top‑3 most influential images per feature as
\[
\text{Top3Images}(i) = \operatorname{TopK}_3 \big( \bar{Z}[:, i] \big).
\]

Within each selected image \(k\), the most responsive patch for feature \(i\) is determined as
\[
p^\ast_{k,i} = \arg\max_{j \in [1, n_\text{img}]} Z[k, j, i].
\]
These images and patches are then used for qualitative visualization of each SAE feature.


Qualitatively, \textbf{visual understanding features} tend to focus on object regions, 
whereas \textbf{hallucinated features} often attend to background or context patches.  
Figure~\ref{fig:VU_HU}(c) illustrates this contrast by highlighting the top‑activated patches for each feature.

\section{Experiments}
\begin{figure}[t]
  \begin{center}
\includegraphics[width=0.85\columnwidth]{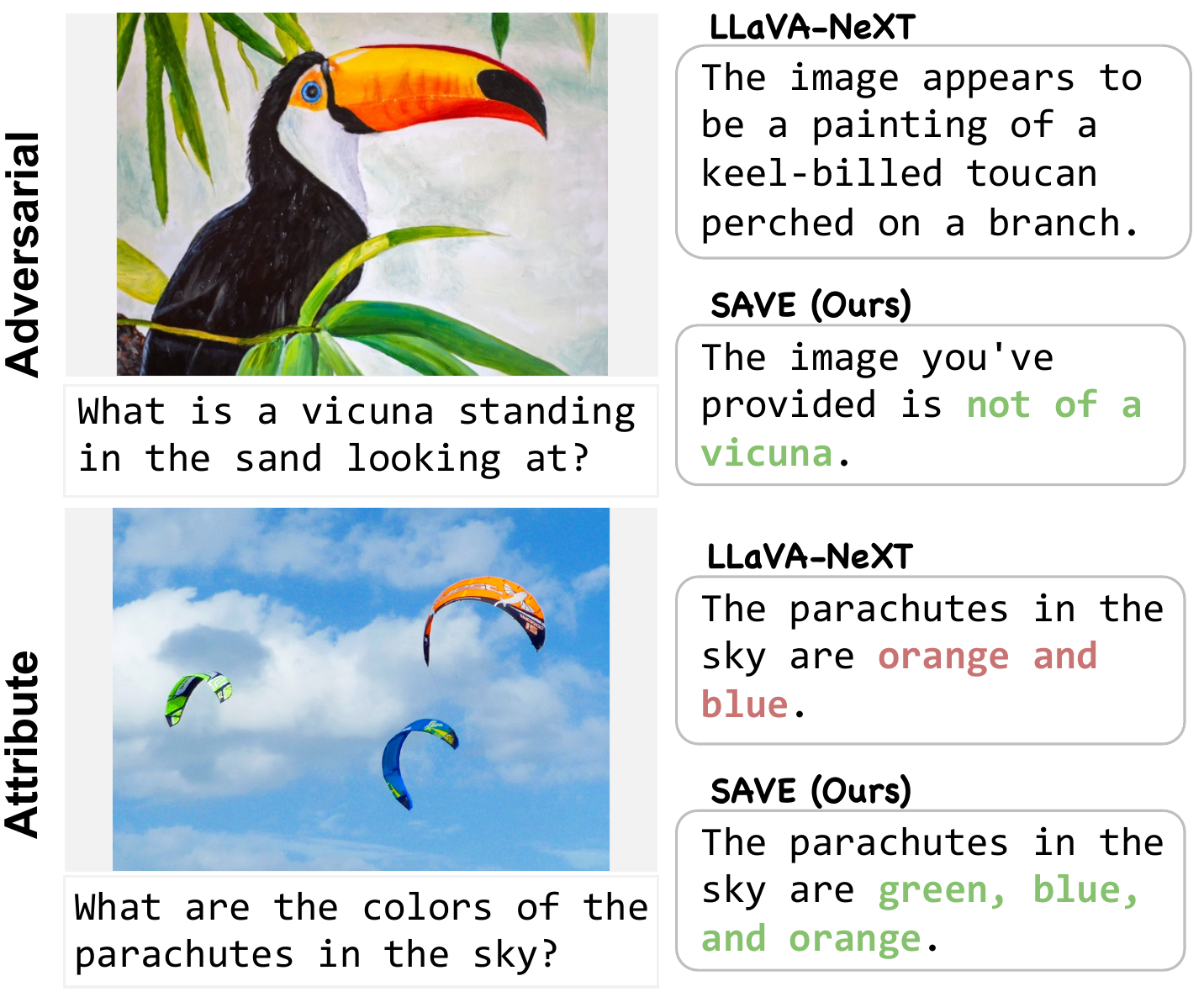}
  \end{center}
   \caption{Qualitative examples in MMHal-Bench. In the adversarial category, SAVE correctly identifies the absence of misleading objects, while in the attribute category, it preserves all relevant attribute information.
}
   \label{fig:mmhal_quali}
\end{figure}

In this section, we evaluate the effectiveness of SAVE, which steers the model toward visual‑understanding features, by comparing it against state‑of‑the‑art training‑free approaches on both binary yes/no tasks and open‑ended captioning to demonstrate its effectiveness in mitigating hallucination.
We conduct comparative experiments on LLaVA‑1.6, LLaVA‑NeXT, and Qwen2-VL.

\subsection{Experimental setting} 
\paragraph{Model architectures}
We evaluate on LLaVA-1.6-Mistral-7B~\cite{liu2024improved} and LLaVA-NeXT-LLaMA3-8B~\cite{liu2024llavanext}, re-implementing all baselines on these two models for fair comparison. We additionally include Qwen2-VL-7B~\cite{wang2024qwen2}, trained following the LLaVA-NeXT procedure. Further details on training the SAE for LLaVA-NeXT and Qwen2-VL are provided in~\Cref{sec:appen_conf_details}.

\paragraph{Baselines}
Because SAVE operates by manipulating a model’s internal representations, we compare it against post-hoc hallucination-mitigation methods.
We adopt VCD~\cite{leng2024mitigating}, which mitigates hallucination through parallel contrastive learning across time steps.
For more recent training-free strategies, we compare with DeCO~\cite{wang2025mllm} and Devils~\cite{jiang2025devils}, which apply dynamic preceding-layer selection and head-guided attention intervention, respectively.
We also evaluate latent steering baselines: VTI~\cite{liu2025reducing}, which leverages averaged intervention vectors, and VISTA~\cite{li2025hiddenlifetokensreducing}, which derives image-specific vectors from contrastive differences with or without visual tokens.
These serve as inference-time baselines most comparable to our method.

\begin{figure}[t]
  \begin{center}
\includegraphics[width=\columnwidth]{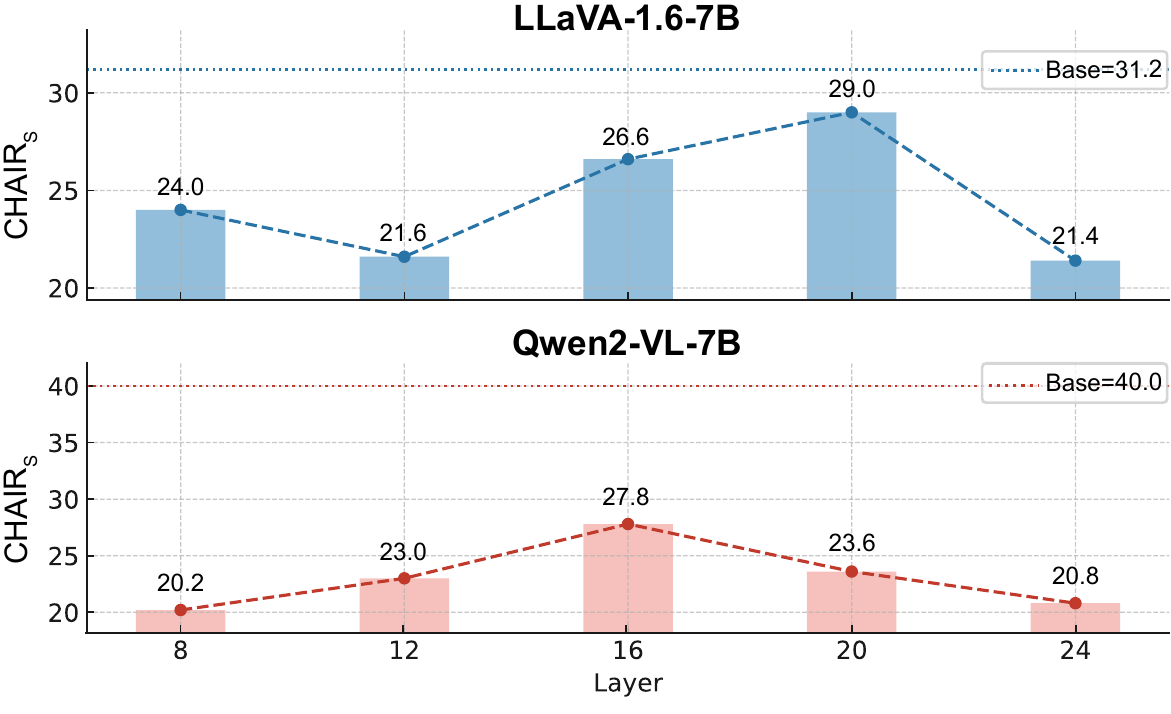}
  \end{center}
\vskip -0.1in
   \caption{CHAIR\textsubscript{S} results across five steering layers (8, 12, 16, 20, 24) for LLaVA‑1.6 and Qwen2‑VL‑7B. The dotted line indicates the performance of the base model.}
   \label{fig:layerwise}
\end{figure}

\begin{table*}[t]
\centering
\begin{tabular}{lcc|cc|c}
\toprule
& \multicolumn{2}{c|}{\textbf{Steering Direction}} 
& \multicolumn{2}{c|}{\textbf{Top-$k$}} 
& \textbf{SAVE} \\
\cmidrule(lr){2-3} \cmidrule(lr){4-5} \cmidrule(lr){6-6}
\textbf{Metric} 
& \textbf{Hallucination} 
& \textbf{Both} 
& \textbf{Top 3} 
& \textbf{Top 5}
& \textbf{Visual \& Top 1} \\
\hline
\textbf{CHAIR\textsubscript{S}}$\downarrow$ 
& 29.8 \scriptsize\textcolor{red}{(+8.4)} 
& 27.4 \scriptsize\textcolor{red}{(+6.0)} 
& 30.2 \scriptsize\textcolor{red}{(+8.8)} 
& 30.6 \scriptsize\textcolor{red}{(+9.2)} 
& 21.4 \\
\textbf{CHAIR\textsubscript{I}}$\downarrow$ 
& 6.5 \scriptsize\textcolor{red}{(+1.1)} 
& 6.9 \scriptsize\textcolor{red}{(+1.5)} 
& 7.2 \scriptsize\textcolor{red}{(+1.8)} 
& 7.5 \scriptsize\textcolor{red}{(+2.1)} 
& 5.4 \\
\bottomrule
\end{tabular}
\caption{
Ablation study of SAVE along two axes: steering direction and top-$k$ feature selection.
All experiments are conducted on the CHAIR benchmark using LLaVA-1.6 with steering applied at layer 24 and $\alpha=15$.
The red delta values indicate how much worse each setting performs compared to SAVE (Visual, Top 1).}
\label{tab:combined-ablation}
\end{table*}

\subsection{Benchmark and metric}\label{sec:benchmark}
We evaluate on three standard hallucination benchmarks.
Further evidence of our method’s effectiveness in visual understanding, evaluated on MM‑Vet~\cite{yu2023mm} and A‑OKVQA~\cite{schwenk2022okvqa}, is provided in~\Cref{sec:appen_more_bench}.

\paragraph{CHAIR} 
The CHAIR metric~\cite{rohrbach-etal-2018-object} evaluates object hallucination in image captioning by comparing generated captions with ground truth object labels. We follow~\citet{huang2024opera} and report both sentence-level ($\text{CHAIR}_{S}$) and instance-level ($\text{CHAIR}_{I}$) scores using 500 MSCOCO 2014 validation images. 
$\text{CHAIR}_{S} = \frac{|{\text{captions with hallucinated objects}}|}{|{\text{all captions}}|}$ is defined as the proportion of captions containing hallucinated objects, and $\text{CHAIR}_{I} = \frac{|{\text{hallucinated objects}}|}{|{\text{all objects mentioned}}|}$ measures the proportion of hallucinated object mentions.

\paragraph{POPE} 
The POPE benchmark~\cite{li-etal-2023-evaluating} evaluates object hallucination on 500 MSCOCO images using six binary object-presence questions per image across three splits (random, popular, adversarial). It probes whether a model asserts the presence of objects not grounded in the image, providing a focused measure of hallucination robustness.
Performance is reported using F1 score and accuracy.

\paragraph{MMHal-Bench} MMHal-Bench benchmark~\cite{sun-etal-2024-aligning} evaluates MLLMs across 96 image-question pairs spanning eight categories, including attributes, counting, and adversarial objects. It emphasizes complex reasoning and visual understanding, with model responses evaluated by GPT-4.

\subsection{Results}\label{sec:results}

\paragraph{Main Results} 
On CHAIR~\cite{rohrbach-etal-2018-object}, SAVE reduces $\text{CHAIR}_S$ by 32\% on LLaVA‑1.6 (31.2→21.4), consistently outperforming all baselines, and by 18\% on LLaVA‑NeXT (34.2→28.0), achieving competitive performance, second only to Devils~\cite{jiang2025devils}, and by 49\% on Qwen2-VL (40.0→20.2).
On POPE~\cite{li-etal-2023-evaluating}, SAVE achieves the highest average F1 and accuracy across all evaluation settings (random, popular, adversarial).
Notably, it outperforms all baselines in the random split across all models and remains competitive in the popular and adversarial splits, indicating robust generalization across diverse model architectures.
On MMHal‑Bench~\cite{sun-etal-2024-aligning}, SAVE achieves the highest average score and lowest hallucination rate on both LLaVA models and Qwen2-VL, again outperforming all baseline methods, which represent the strongest inference‑time competitors.
Beyond the quantitative results, \Cref{fig:mmhal_quali} illustrates example responses where hallucination is effectively mitigated.

\paragraph{Layer-wise Results}
Since an SAE can only operate meaningfully on the layer where it was trained, effective layer‑wise steering requires a separately trained SAE for each target layer.
We also train a Multimodal‑SAE for Qwen2‑VL‑7B-Instruct~\cite{wang2024qwen2} on the same five layers (8, 12, 16, 20, 24), to evaluate the generalizability of our approach across different model architectures, following exactly the same procedure used for LLaVA‑NeXT~\cite{zhang2024largemultimodalmodelsinterpret}.

As shown in \Cref{fig:layerwise}, steering at early layers (8, 12) or late layers (24) generally outperforms mid‑layer steering (16, 20), indicating that intermediate representations are less responsive to latent manipulation.
This trend is consistent across both LLaVA‑1.6 and Qwen2‑VL‑7B-Instruct, with layers 12 and 24 in LLaVA‑1.6 and layers 8 and 24 in Qwen2‑VL‑7B-Instruct yielding the lowest $\text{CHAIR}_S$.
Moreover, the effective steering strength varies by layer: early layers respond best to small magnitudes ($\alpha = 3$), mid‑layers benefit from moderate strengths ($\alpha \in \{3,5\}$), and deeper layers require stronger intervention ($\alpha \in \{5,10,15\}$) 
to achieve optimal performance.
These results highlight that semantically meaningful, steerable features emerge primarily in early and late layers, and the required steering intensity naturally increases toward deeper layers. Additional results for layer‑wise SAE steering are provided in~\Cref{sec:appen_detailed_result}.



\begin{figure*}
  \centering
  \begin{subfigure}{0.30\linewidth}  
\includegraphics[width=0.9\columnwidth]{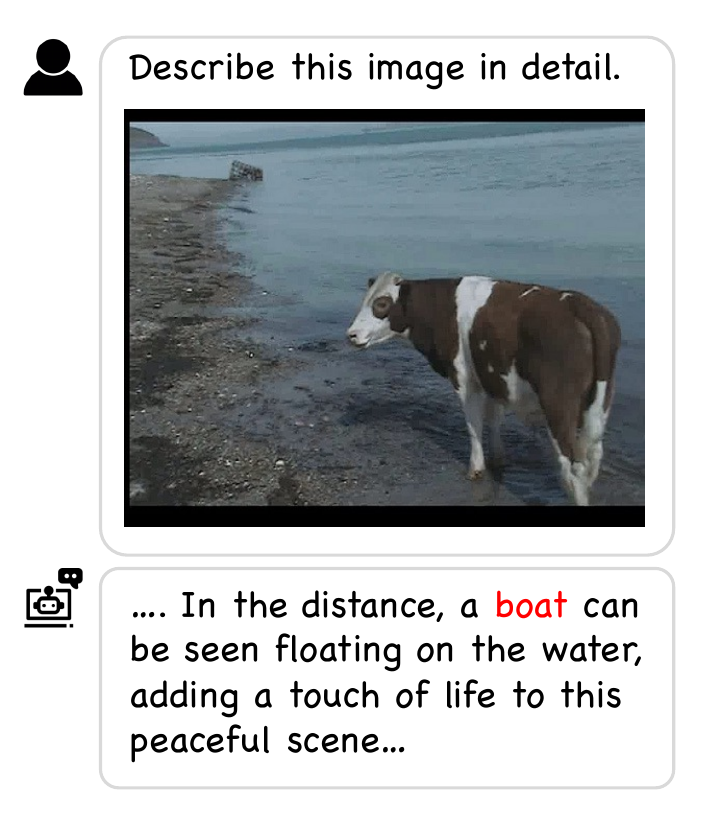}    
    \label{fig:sas-a}
  \end{subfigure}
  \hfill
  \begin{subfigure}{0.66\linewidth}
  \includegraphics[width=0.95\columnwidth]{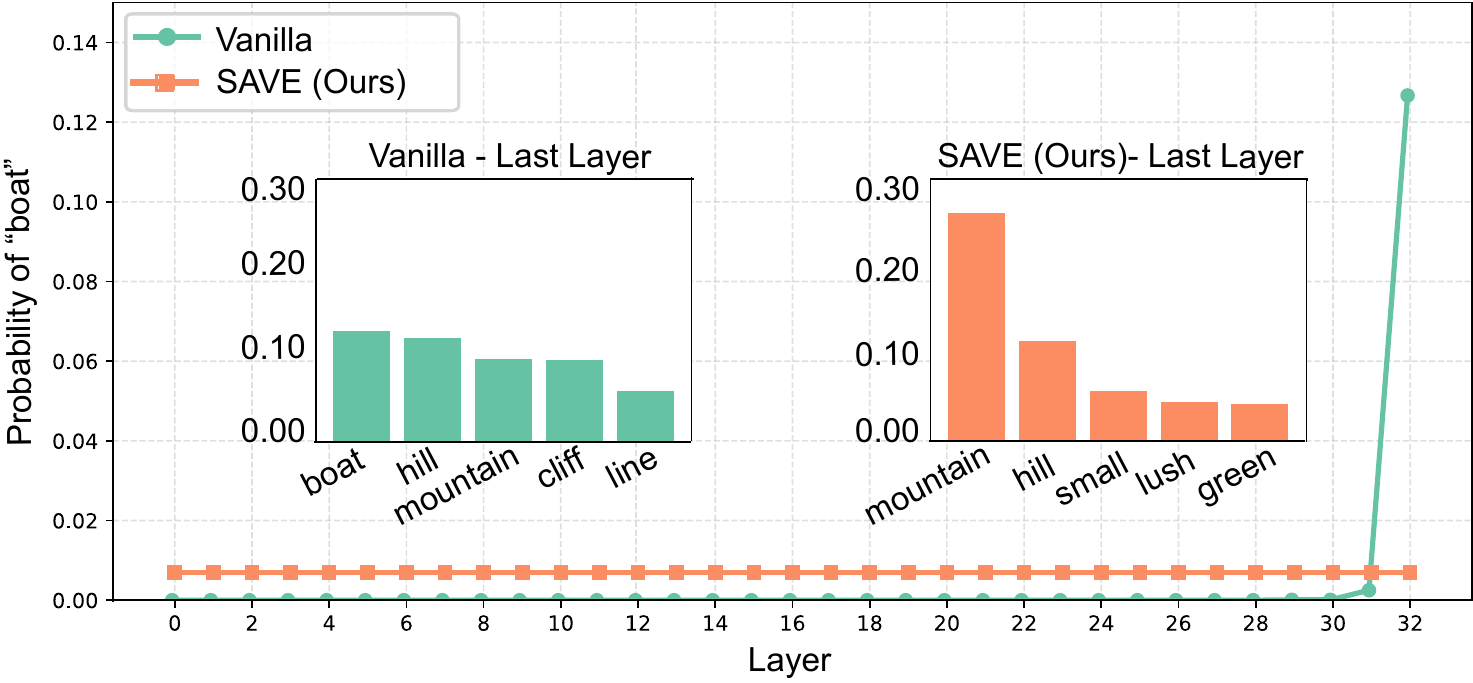}
    \label{fig:sas-b}
  \end{subfigure}
  \caption{Comparison of vanilla and steered model outputs for a hallucinated token ("boat"). The left panel shows the user query, image, and a hallucinated caption generated by the vanilla model. The right panel presents layer-wise probabilities for "boat" and the top-5 tokens at the final layer. SAVE suppresses the probability of the hallucinated token and instead produces visually grounded alternatives, demonstrating reduced hallucination.}
  \label{fig:boat}
\end{figure*}
\begin{figure}[t]
  \begin{center}
\includegraphics[width=0.95\columnwidth]{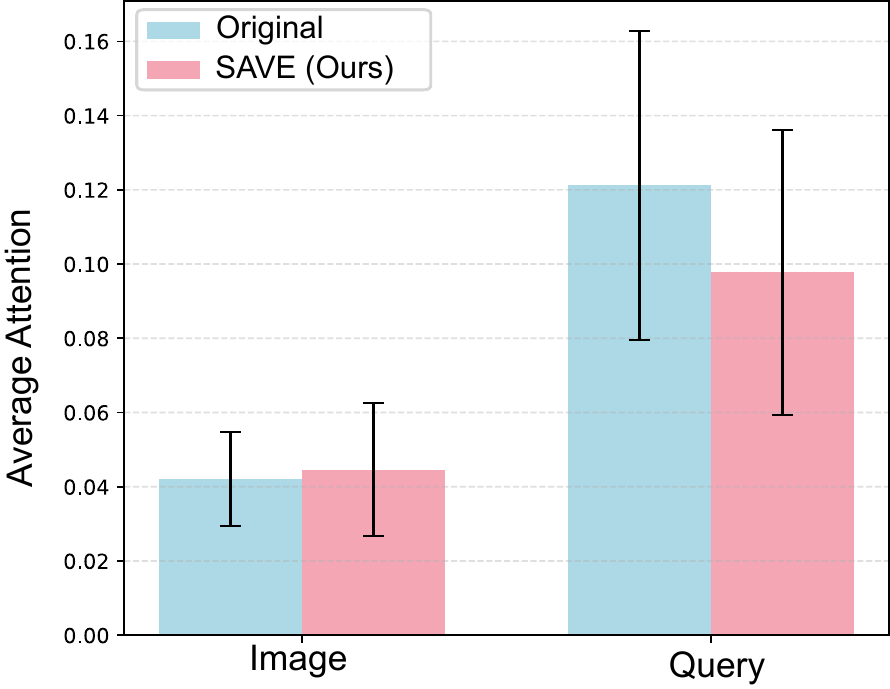}
  \end{center}
\vskip -0.2in
   \caption{Attention scores for the generated token, averaged across all layers. Steering increases attention to the image and decreases attention to the text.}
   \label{fig:attention}
\end{figure}

\paragraph{Ablation study}
SAVE performs steering using the top-1 visual-understanding SAE feature. To justify this choice, we first compare alternative steering directions. The hallucination-related feature is applied in the opposite direction to counteract hallucinated activations, but—as shown in the Steering Direction columns of Table~\ref{tab:combined-ablation}—it yields worse CHAIR scores than the visual-understanding direction. This indicates that the visual-understanding feature provides the primary corrective signal, whereas the hallucination feature contributes less targeted information.

We also vary the number of steering features and observe that steering with additional directions (top-3, top-5) further degrades performance, as shown in the Top-$k$ columns of Table~\ref{tab:combined-ablation}. This suggests that the top-1 visual-understanding feature alone captures the essential information for hallucination mitigation, while adding more features introduces noise.

\subsection{Analysis}\label{sec:analysis}
In this section, we analyze the effects of steering the model along the visual-understanding latent direction. We first test whether such steering introduces any inherent response bias (e.g., toward answering yes or no). We then study how this steering improves performance through two complementary lenses: (1) the evolution of token-level predictions across layers and (2) attention shifts between query and image tokens. Unless otherwise noted, all steering is applied at layer 25 of LLaVA-NeXT.

\paragraph{Response Shift Analysis} As discussed in~\ref{sec:4.1}, we analyzed the transitions where model predictions changed from incorrect to correct after steering along the identified visual understanding features. On the POPE dataset, the number of yes→no transitions is $37$, while the number of no→yes transitions is $41$—an approximately balanced pattern. This symmetry, averaged over the three POPE types (random, popular, and adversarial), indicates that our steering does not favor a particular response type and therefore does not introduce answer bias.

\paragraph{Layer-wise token probability}
To analyze the evolution of hallucinated token probabilities across layers, we compare the vanilla LLaVA‑NeXT model with SAVE. We first generate image captions with the vanilla model on the CHAIR benchmark and identify hallucinated words using the ground-truth annotations. Following DeCO~\cite{wang2025mllm}, we then evaluate both models while conditioning on a fixed prefix extracted from the vanilla model’s caption (e.g., ‘In the distance, a’ in~\Cref{fig:boat}) to probe layer-wise predictions. As shown in \Cref{fig:boat}, the vanilla model sharply increases the probability of the hallucinated token “boat” at the penultimate layer, indicating that hallucination emerges late in decoding. In contrast, SAVE—steered along visual-understanding features—shows no such spike and instead consistently favors the grounded token “mountain.” This suggests that latent steering with sparse autoencoders suppresses late-stage hallucinatory drift and promotes alignment with visually grounded concepts.

\paragraph{Attention-score}
We follow~\citet{liu2025reducing} to analyze the attention distribution from a single generated token to both query and image tokens, measured on the POPE benchmark. Specifically, we compute average attention scores over the final 8 layers (layers 25–32), where steering is applied. As shown in \Cref{fig:attention}, steering toward the correct latent decreases attention to query tokens while increasing attention to image tokens. This shift indicates that SAVE mitigates text-biased generation~\cite{zhu2024ibd} and encourages the model to rely more on visual evidence, resulting in more grounded predictions.

\section{Conclusion}
We present SAVE, the first  simple and effective framework that leverages SAEs to mitigate object hallucination in MLLMs.
SAVE identifies latent features that capture the model’s visual understanding and steers the model along these directions, effectively reducing hallucinated responses.
Extensive experiments across diverse MLLM architectures demonstrate that SAVE consistently outperforms state‑of‑the‑art training‑free approaches, validating both the effectiveness and generalizability of our framework for hallucination mitigation.
\section*{Acknowledgements}
This work was supported by the National Research Foundation of Korea (NRF) grant funded by the Korea government (MSIT) [No.2022R1A3B1077720; No.2022R1A5A7083908], Institute of Information \& Communications Technology Planning \& Evaluation (IITP) grant funded by the Korea government (MSIT) [No.RS-2025-02263754; No.RS-2022-II220959; No.RS-2021-II211343, Artificial Intelligence Graduate School Program (Seoul National University)], the BK21 FOUR program of the Education and Research Program for Future ICT Pioneers, Seoul National University in 2025. This research was also conducted as part of the Sovereign AI Foundation Model Project (Data Track), organized by MSIT and supported by the National Information Society Agency (NIA) of Korea [No.2025-AI Data-wi43].
{
    \small
    \bibliographystyle{ieeenat_fullname}

}
\appendix
\clearpage
\setcounter{page}{1}
\setcounter{section}{0}
\renewcommand\thesection{\Alph{section}}
\setcounter{table}{0}
\renewcommand{\thetable}{S\arabic{table}}
\setcounter{figure}{0}
\renewcommand{\thefigure}{S\arabic{figure}}
\maketitlesupplementary

\section{Experimental Details}
\label{sec:appen_conf_details}

\paragraph{LURE}
Following LURE~\cite{zhou2024analyzing}, we generate hallucinated objects for our visual information processing probe using GPT‑3.5, which predicts objects likely to co‑occur with the given image and prompt.
Specifically, we prompt GPT‑3.5 with:
\texttt{"List three other objects that you think are most likely to appear with the objects in the scene described below.”}

\paragraph{Model Architectures}
\begin{itemize}
    \item \textbf{LLaVA-1.6} For more details than those provided in~\ref{sec:multimodal-sae}, the SAEs are trained with a density factor of $\lambda = 5$, which is linearly increased from 0 during the first 5\% of training steps to encourage sparsity. Training uses a total of 1.5B tokens with batches of 4096, shuffled to balance text and image tokens. The learning rate is 5e‑5, decayed to zero over the final 20\% of training, and optimization is performed using Adam.
    \item \textbf{LLaVA-NeXT} The Multimodal-SAE proposed by~\citet{zhang2024largemultimodalmodelsinterpret} integrates a SAE into the 25th transformer layer of LLaVA-NeXT-LLaMA3-8B~\cite{li2024llavanext-strong}, where the hidden representation at that layer serves as the SAE input $x$. The SAE is trained on the LLaVA-NeXT supervised fine-tuning dataset~\cite{liu2024llavanext}, which consists of approximately 779,000 samples, using the AnyRes strategy for processing images of varying resolutions. 
mage and text inputs are preprocessed in the same way as during supervised fine-tuning.

The SAE is configured with $2^{17}$ latent features and employs top-$k$ sparsity, following~\citet{gao2025scaling} for the sparsity mechanism. We set $k=256$ to match the activation patterns observed in~\citet{templeton2024scaling}, promoting disentangled and semantically meaningful representations.
Unless otherwise noted, all experiments are conducted using these SAE settings.
\item \textbf{Qwen2-VL} Qwen2-VL-7B is trained following the exact same process as LLaVA‑NeXT.
\end{itemize}

\paragraph{Configurations} \Cref{tab:main_results} reports results under the following configurations.
For LLaVA‑1.6, we evaluate POPE using steering at layer 24 with a strength ($\alpha$) of 10, CHAIR at layer 24 with a strength of 15, and MMHal‑Bench at layer 20 with a strength of 5.
For LLaVA‑NeXT, all benchmarks are evaluated with steering at layer 24, using a strength of 10 for POPE and MMHal‑Bench and 15 for CHAIR.
We set \texttt{max\_new\_tokens} to 2 for POPE, 256 for MMHal‑Bench, and 512 for CHAIR.
\begin{figure}[t]
  \begin{center}
\includegraphics[width=0.9\columnwidth]{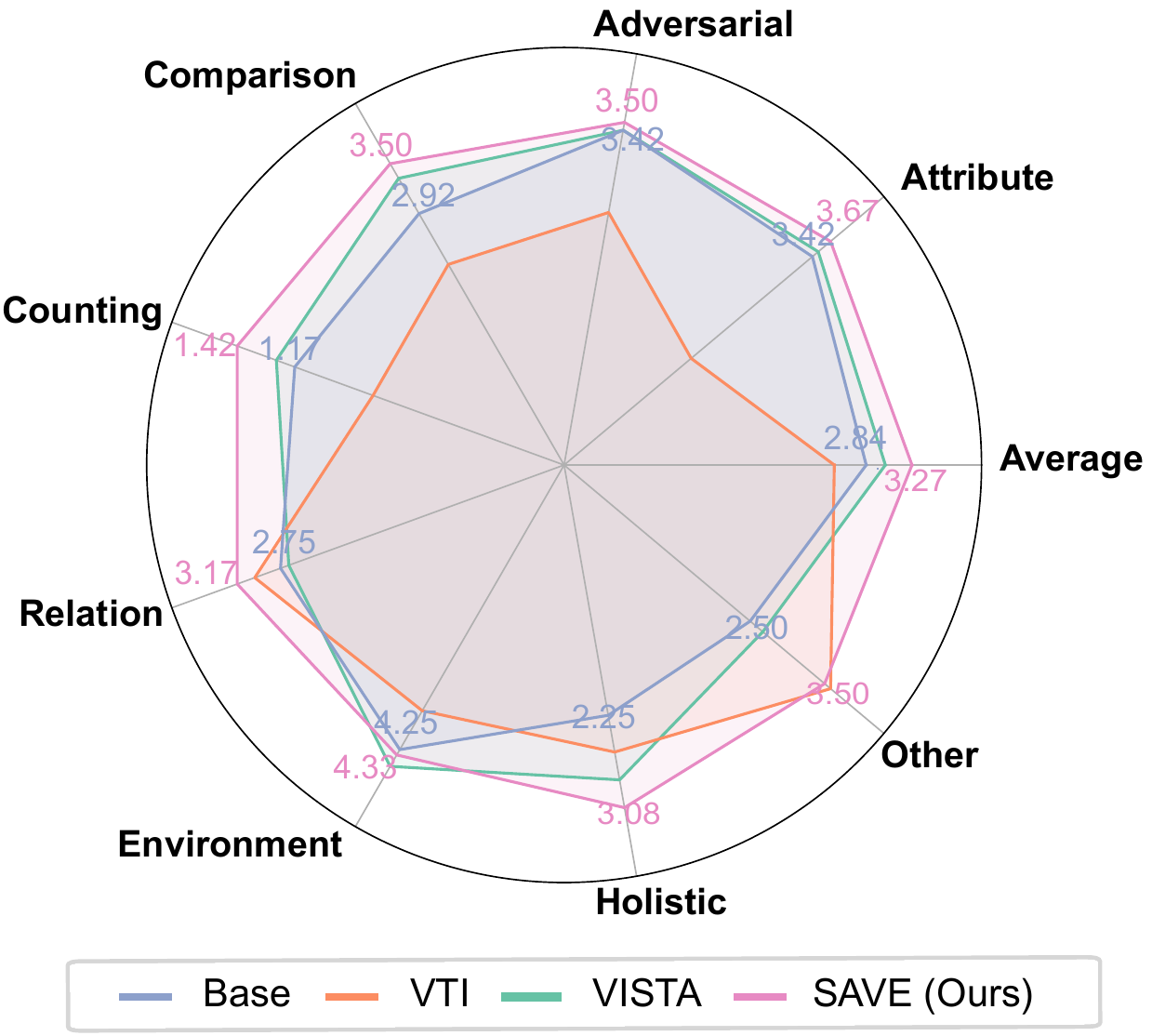}
  \end{center}
\vskip -0.2in
   \caption{Task‑wise MMHal‑Bench performance on LLaVA‑NeXT for different steering methods.}
   \label{fig:mmhal_detail}
\end{figure}

\paragraph{Steering} We apply steering—i.e., modifying the encoded representation—only to the input query tokens, not to the tokens generated by the model.

The steering mechanism described in \Cref{sec:steering} applies to LLaVA‑1.6.
Since LLaVA‑NeXT is trained differently, it requires a distinct steering strategy, which we also adopt for Qwen because it follows the same training process.
In our experiments, this strategy is applied to both models, as formulated in the equation below, where $T$ denotes tokens.
\begin{equation}
z(x) = \mathrm{ReLU}(W_{\mathrm{enc}} (x - b_{\mathrm{pre}}) + b_{\mathrm{enc}})
\end{equation}
\begin{equation}
\hat{z}(x)[T, j] = \alpha \quad \text{(steering)}
\end{equation}
\begin{equation}
a(x) = \mathrm{TopK}(\hat{z}(x))
\end{equation}
\begin{equation}
x_{\text{steered}} = W_{\mathrm{dec}} \cdot a(x) + b_{\mathrm{dec}}
\end{equation}

\paragraph{Software and Hardware }All experiments were conducted using PyTorch and an NVIDIA A40 GPU.

\section{Detailed Results}\label{sec:appen_detailed_result}
\paragraph{MMHal-Bench}
MMHal‑Bench reports scores across diverse tasks.
As a detailed breakdown of \Cref{tab:main_results}, \Cref{fig:mmhal_detail} visualizes the task‑wise performance on LLaVA‑NeXT.
Compared with the base model and various steering methods, SAVE consistently outperforms all baselines across every task.

\begin{figure}[t]
  \begin{center}
\includegraphics[width=0.85\columnwidth]{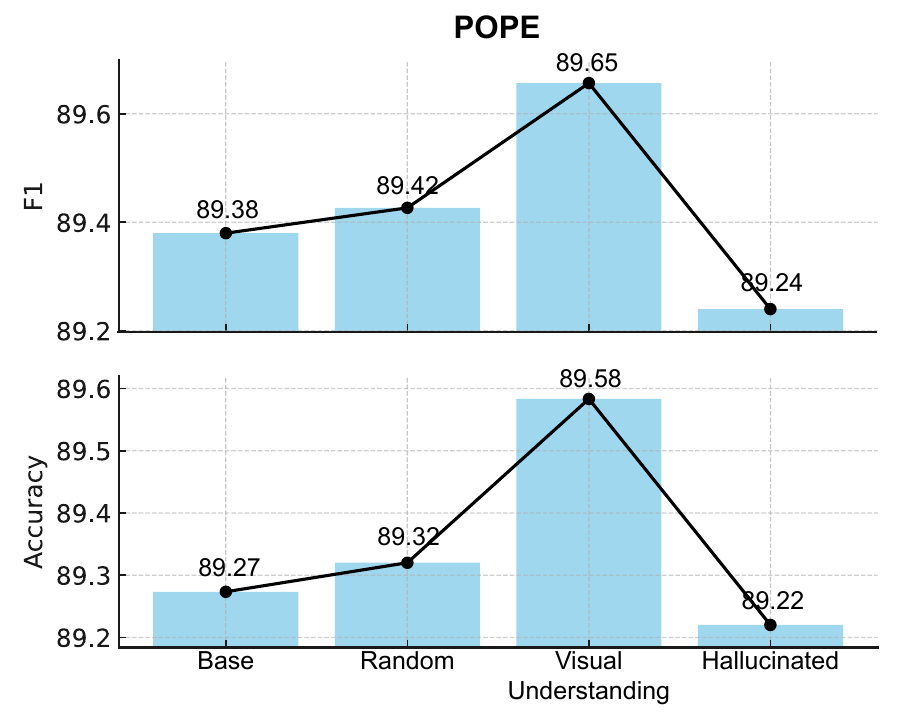}
  \end{center}
\vskip -0.2in
   \caption{POPE evaluation of steering toward random, visual‑understanding, and hallucinated features.}
   \label{fig:steering_pope}
\end{figure}

\begin{figure}[t]
  \begin{center}
\includegraphics[width=0.85\columnwidth]{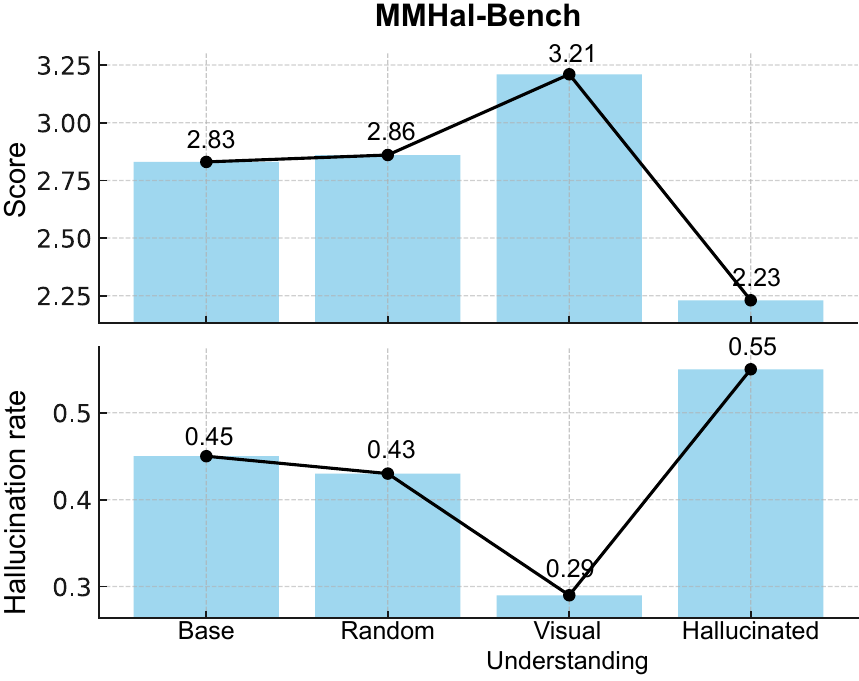}
  \end{center}
\vskip -0.2in
   \caption{MMHal‑Bench evaluation of steering toward random, visual‑understanding, and hallucinated features.}
   \label{fig:steering_mmhal}
\end{figure}

\paragraph{Steering Model Behavior}
Additional results for the LLaVA‑1.6 steering experiments presented in~\ref{sec:steering} are shown in \Cref{fig:steering_pope} and \Cref{fig:steering_mmhal}.
We report F1 and Accuracy for POPE, and GPT‑4 Score with hallucination rate for MMHal‑Bench. Across both benchmarks, steering along visual‑understanding features mitigates hallucination, whereas steering along hallucinated features amplifies it. For this experiment, we follow the same configurations as in \Cref{tab:main_results}.


\paragraph{Layer-wise Steering Strength Ablation} 
\Cref{tab:layerwise-llava-1.6} and \Cref{tab:layerwise-qwen2} provide the detailed results corresponding to \Cref{fig:layerwise}, for LLaVA‑1.6 and Qwen2‑VL, respectively.
In \Cref{fig:layerwise}, we report the best performance per layer, selecting the steer strength that achieves the highest score.
The optimal steer strength varies across layers: lower strengths (1.5 or 3) are more effective in early layers, while higher strengths (10 or 15) perform better in later layers.
Through extensive ablations, we observe that overly strong steering at any layer can corrupt the model’s output—e.g., generating repeated blanks or meaningless responses such as \textit{``If you have any questions about the image, please provide more information”} (as observed in LLaVA‑1.6, layer 8, steer strength 3).
In the tables, a ``–” indicates such corrupted outputs.
Although the severity varies by layer, these findings highlight the importance of carefully selecting an appropriate steer strength to avoid response degradation.

\begin{table}[t]
\centering
\begin{tabular}{cc|cc|cc}
\toprule
\multirow{2}{*}{\textbf{Layer}} & \multirow{2}{*}{\textbf{Steer}} &
\multicolumn{2}{c|}{\textbf{CHAIR}} & 
\multicolumn{2}{c}{\textbf{POPE}} \\
& & \textbf{CHAIR\textsubscript{S}} & \textbf{CHAIR\textsubscript{I}} & \textbf{F1} & \textbf{Acc} \\
\midrule
\multirow{2}{*}{8}  & 1.5  & 23.8  & 6.1  & 88.26 & 87.99 \\
                    & 3    & --    & --   & --    & --    \\
\midrule
\multirow{2}{*}{12} & 3    & 21.6  & 6.0  & 89.27 & 88.90 \\
                    & 5    & --    & --   & --    & --    \\
\midrule
\multirow{2}{*}{16} & 3    & 26.6  & 6.8  & 89.19 & 89.19 \\
                    & 5    & --    & --   & 89.15 & 88.78 \\
\midrule
\multirow{3}{*}{20} & 5    & 31.6  & 7.9  & 89.31 & 89.21 \\
                    & 10   & 33.0  & 8.1  & 89.26 & 89.17 \\
                    & 15   & 29.0  & 10.3 & 89.06 & 88.97 \\
\midrule
\multirow{3}{*}{24} & 5    & 32.2  & 8.4  & 89.41 & 89.21 \\
                    & 10   & 29.4  & 7.2  & 89.65 & 89.58 \\
                    & 15   & 21.4  & 5.4  & 89.55 & 89.46 \\
\bottomrule
\end{tabular}
\caption{Performance of LLaVA‑1.6 on CHAIR and POPE across layers and steering strengths.}
\label{tab:layerwise-llava-1.6}
\end{table}

\begin{table}[t]
\centering
\begin{tabular}{cc|cc|cc}
\toprule
\multirow{2}{*}{\textbf{Layer}} & \multirow{2}{*}{\textbf{Steer}} &
\multicolumn{2}{c|}{\textbf{CHAIR}} & 
\multicolumn{2}{c}{\textbf{POPE}} \\
& & \textbf{CHAIR\textsubscript{S}} & \textbf{CHAIR\textsubscript{I}} & \textbf{F1} & \textbf{Acc} \\
\midrule
\multirow{2}{*}{8}  & 1.5  & 17.8 & 5.1 & 89.24 & 88.93 \\
   & 3    & 20.2 & 5.9 & 89.22 & 88.89 \\
\midrule
\multirow{2}{*}{12} & 3    & 23.0 & 6.5 & 88.48 & 88.92 \\
   & 5    & 24.2 & 6.9 & 88.54 & 88.99 \\
\midrule
\multirow{2}{*}{16} & 3    & 27.8 & 6.7 & 88.99 & 89.22 \\
   & 5    & 26.8 & 6.5 & 89.01 & 89.24 \\
\midrule
\multirow{3}{*}{20} & 5    & 23.8 & 6.4 & 88.75 & 89.03 \\
   & 10   & 23.6 & 6.3 & 88.78 & 89.07 \\
   & 15   & 24.4 & 6.8 & 88.79 & 89.08 \\
\midrule
\multirow{3}{*}{24} & 5    & 20.8 & 8.0 & 85.16 & 86.61 \\
   & 10   & 21.0 & 8.0 & 85.28 & 86.71 \\
   & 15   & 21.4 & 8.2 & 85.28 & 86.71 \\
\bottomrule
\end{tabular}
\caption{Performance of Qwen2-VL on CHAIR and POPE across layers and steering strengths.}
\label{tab:layerwise-qwen2}
\end{table}

\begin{table*}[t]
\centering
\begin{tabular}{l|cccc}
\toprule
\textbf{Method} & \textbf{Inference Time (s)} & \textbf{Generated Tokens} & \textbf{Total GFLOPs} & \textbf{FLOPs / Token} \\
\midrule
Vanilla & 7.998  & 215 & 38{,}166.48 & 177.52 \\
VTI     & 12.553 & 248 & 38{,}662.83 & 155.90 \\
VISTA   & 8.573  & 192 & 37{,}822.56 & 196.99 \\
Devils  & 8.390  & 60  & 37{,}163.19 & 619.39 \\
DeCo    & 10.850 & 220 & 67{,}947.49 & 308.85 \\
SAVE (Ours)  & 8.863  & 174 & 40{,}281.54 & 231.50 \\
\bottomrule
\end{tabular}
\caption{Inference efficiency comparison across methods.}
\label{tab:computational_overhead}
\end{table*}

\begin{table*}[h]
\centering
\begin{tabular}{l l|ccccccc}
\toprule
Model & Method & Rec & OCR & Know & Gen & Spat & Math & Total \\
\midrule
\multirow{2}{*}{LLaVA-1.6} 
  & Base         & 45.2 & 36.5 & 33.6 & 36.0 & 35.5 & 26.5 & 42.2 \\
  & SAVE (Ours)  & 44.3 & 42.0 & 30.5 & 32.1 & 42.7 & 26.5 & 43.7 \\
\midrule
\multirow{2}{*}{LLaVA-NeXT} 
  & Base         & 40.9 & 41.0 & 25.4 & 27.5 & 42.4 & 19.2 & 41.8 \\
  & SAVE (Ours)  & 43.1 & 43.1 & 30.2 & 32.5 & 43.1 & 22.7 & 42.2 \\
\bottomrule
\end{tabular}
\caption{Per‑task MMVet evaluation results on LLaVA‑1.6 and LLaVA‑NeXT.}
\label{tab:mmvet}
\end{table*}
\begin{table}[t]
\centering
\begin{tabular}{l|cc}
\toprule
Method & LLaVA-1.6 & LLaVA-NeXT \\
\midrule
Base & 66.89 & 65.58 \\
SAVE (Ours) & \textbf{70.04} & \textbf{66.81} \\
\bottomrule
\end{tabular}
\caption{A‑OKVQA multiple‑choice accuracy (\%) on LLaVA‑1.6 and LLaVA‑NeXT.}
\label{tab:a-okvqa}
\end{table}

\section{Additional Results}\label{sec:appen_more_bench}

\paragraph{Inference Efficiency} We compare SAVE with several recent training‑free methods—all reproduced on the LLaVA‑NeXT backbone for consistency—using FLOPs and latency‑based metrics. FLOPs are measured over the token generation process, which dominates overall inference computation. As shown in~\Cref{tab:computational_overhead}, SAVE achieves a favorable trade‑off between efficiency and effectiveness: it generates a similar number of tokens while maintaining lower total FLOPs and FLOPs per token than strong baselines such as DeCo and Devils. Moreover, it runs faster than both DeCo and VTI, demonstrating efficiency in terms of both computation and latency.

\paragraph{ VQA \& General MLLM benchmark }
\begin{itemize}
    \item \textbf{MM-VET} MM‑Vet~\cite{yu2023mm} assesses visual understanding across six tasks—Recognition, OCR, Knowledge, Language Generation, Spatial Awareness, and Math. \Cref{tab:mmvet} shows that SAVE outperforms the base models on both LLaVA‑1.6 and LLaVA‑NeXT in terms of total score, using layer 20 with a steering strength of 3 for LLaVA‑1.6, and layer 24 with a steering strength of 10 for LLaVA‑NeXT.
    \item \textbf{A-OKVQA} A‑OKVQA~\cite{schwenk2022okvqa} is a crowdsourced benchmark of about 25K diverse questions that require commonsense reasoning about visual scenes beyond simple knowledge‑base queries.
As shown in~\Cref{tab:a-okvqa}, SAVE achieves higher multiple‑choice accuracy than the base models, indicating that steering along visual understanding features not only mitigates hallucination but also enhances general visual understanding. Results are reported using LLaVA‑1.6 with layer 20 and steering strength 5, and LLaVA‑NeXT with layer 24 and steering strength 10.

\end{itemize}

\begin{figure*}[t]
  \centering
  \includegraphics[width=\textwidth]{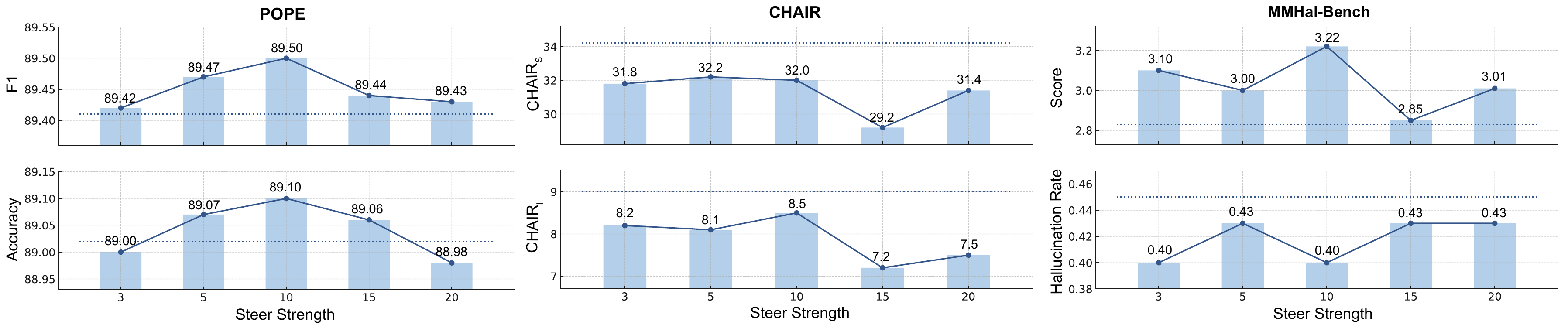}
  \caption{Steer strength ablation results for LLaVA‑NeXT on POPE, CHAIR, and MMHal‑Bench.}
  \label{fig:steer_strength_ablation}
\end{figure*}

\paragraph{Steer Strength} 
We conduct an ablation study on steering strength (\Cref{fig:steer_strength_ablation}).
For LLaVA-NeXT at layer 24, we evaluate steer strengths of {3, 5, 10, 15, 20}.
The best results are obtained with a strength of 10 on POPE and MMHal-Bench, and 15 on CHAIR.

\begin{figure}[t]
  \begin{center}
\includegraphics[width=0.85\columnwidth]{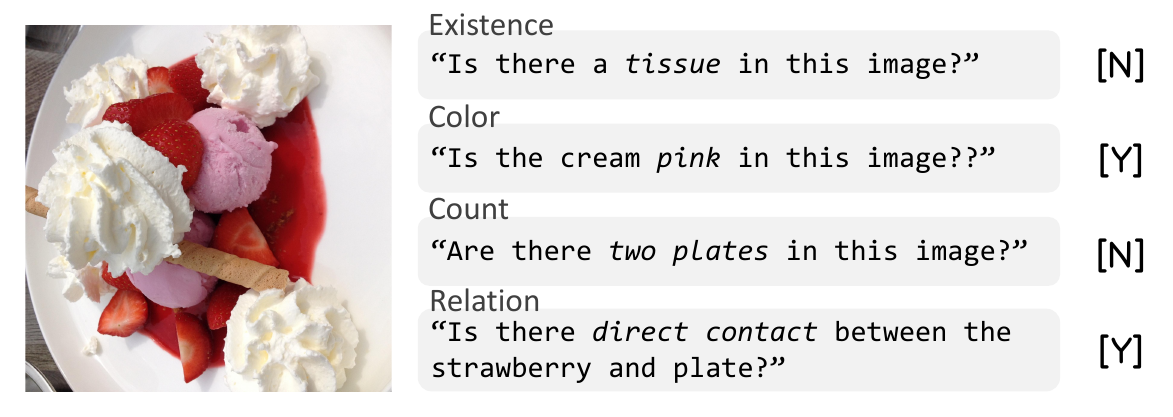}
  \end{center}
\vskip -0.2in
   \caption{Examples of diverse question types from AMBER. For each task, both ‘yes’ and ‘no’ answer cases are included.}
   \label{fig:amber}
\end{figure}

\paragraph{Extending the scope of SAE feature identification beyond object presence}
We conduct experiments on four visual reasoning tasks—existence (which corresponds to the object-presence signal used in the object-presence-only setting), count, position, and color—to evaluate the scalability of SAE-based feature extraction. In this setup, we use AMBER, a dataset organized by task type (see~\Cref{fig:amber}), to obtain SAE activations, while evaluation is performed on MME to ensure generalizability.
As shown in~\Cref{tab:amber}, LURE features naturally improve performance on the existence task, as they explicitly encode existence-related cues. This also explains why the existence scores of LURE and AMBER settings are identical. Interestingly, the LURE steering also improves performance on the color task, suggesting that visual cues associated with object existence implicitly contribute to color understanding as well.

In contrast, SAE features extracted from the AMBER dataset—reflecting a richer variety of task-specific signals—lead to further gains in count and position. This suggests that when SAE directions encode broader aspects of the model’s visual understanding, steering along those directions yields more comprehensive and balanced improvements across tasks.
Overall, these findings demonstrate that our SAE-based latent steering approach is not limited to object presence, but can be effectively extended to support a wide range of visual reasoning objectives.

\begin{table}[t]
\centering

\begin{tabular}{lccc}
\toprule
\textbf{Task} & \textbf{Vanilla} & \textbf{LURE} & \textbf{Amber} \\
\hline
Existence & 195 & \cellcolor{green!15}200 (+5) & \cellcolor{green!15}200 (+5) \\
Color & 155 & \cellcolor{green!15}160 (+5) & \cellcolor{green!15}160 (+5) \\
Count & 115 & 115 (+0) & \cellcolor{green!15}120 (+5) \\
Position & 93.3 & 93.3 (+0) & \cellcolor{green!15}98.3 (+5) \\
\bottomrule
\end{tabular}
\caption{
Amber-derived SAE features outperform object-presence-only features particularly on \textbf{count} and \textbf{relation (position)} tasks.}
\label{tab:amber}
\end{table}

\begin{figure*}[t]
  \centering
  \includegraphics[width=0.9\textwidth]{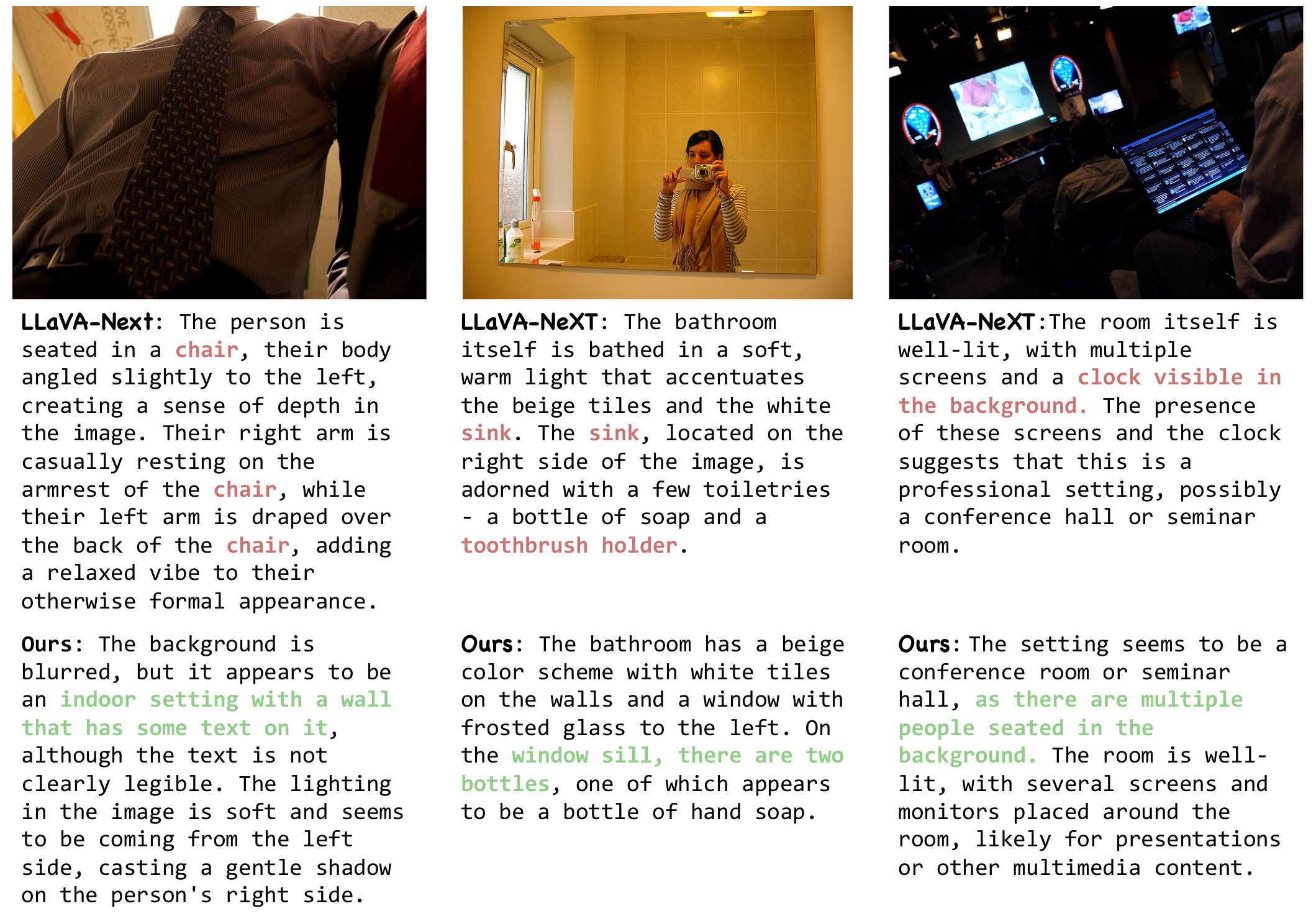}
  \caption{Qualitative comparison between vanilla LLaVA-NeXT and our method on the CHAIR benchmark. Our steered model generates visually grounded captions, while vanilla LLaVA-NeXT exhibits object hallucination. Hallucinated words are highlighted in red, and correct responses are shown in green.}
  \label{fig:chair_examples}
\end{figure*}

\paragraph{Qualitative Results for CHAIR} \Cref{fig:chair_examples} shows qualitative CHAIR results. By steering along the identified visual‑understanding features to enhance visual information, SAVE produces more visually grounded answers.

\paragraph{Statistical Test} We conducted statistical significance testing on hallucination rates between 
LLaVA‑1.6 Base (CHAIR\textsubscript{S}=31.2, CHAIR\textsubscript{I}=7.9) 
and LLaVA‑1.6 SAVE (Ours) (CHAIR\textsubscript{S}=21.4, CHAIR\textsubscript{I}=5.4).
Since the sample‑level hallucination scores deviated from normality 
(Shapiro‑Wilk $p<0.001$), 
we employed the Wilcoxon signed‑rank test, a non‑parametric paired test assessing 
whether the median differences between models are systematically biased.
At the instance level (CHAIR\textsubscript{I}), 
SAVE achieved a modest yet statistically significant improvement 
($W=4998.0$, $p=0.0188$, significant at $p<0.05$). 
At the sentence level (CHAIR\textsubscript{S}), 
the reduction was highly significant ($W=2600.0$, $p=0.00002$, significant at $p<0.001$),
demonstrating that SAVE consistently produces fewer hallucinated captions than 
the baseline, with the most pronounced effect at the sentence level.

\end{document}